\documentclass{article} 
\usepackage{iclr2017_conference,times}
\usepackage{hyperref}
\usepackage{url}
\usepackage[cmex10]{amsmath}
\usepackage{graphicx,multirow,array}
\usepackage{algorithm}
\usepackage{algpseudocode}
\usepackage{caption}
\usepackage{subcaption}
\usepackage{paralist}
\usepackage[title,titletoc,toc]{appendix}
\usepackage[flushleft]{threeparttable} 

\graphicspath{{./}{./figures/}}

\algdef{SE}[DOWHILE]{Do}{doWhile}{\algorithmicdo}[1]{\algorithmicwhile\ #1}%

\title{Alternating Direction Method of Multipliers for Sparse Convolutional Neural Networks}

\author{Farkhondeh Kiaee, Christian Gagn\'{e}, and Mahdieh Abbasi\\
Computer Vision and Systems Laboratory\\
Department of Electrical Engineering and Computer Engineering\\
Universit\'e Laval, Qu\'ebec, QC~~G1V 0A6, Canada \\
\texttt{\{farkhondeh.kiaee.1,mahdieh.abbasi.1\}@ulaval.ca} \\\texttt{christian.gagne@gel.ulaval.ca}
}

\newcommand{\KernelLetter}{W}

\begin{document}

\maketitle
\begin{abstract}
The storage and computation requirements of Convolutional Neural Networks (CNNs) can be prohibitive for exploiting these models over low-power or embedded devices. This paper reduces the computational complexity of the CNNs by minimizing an objective function, including the recognition loss that is augmented with a sparsity-promoting penalty term. The sparsity structure of the network is identified using the Alternating Direction Method of Multipliers (ADMM), which is widely used in large optimization problems. This method alternates between promoting the sparsity of the network and optimizing the recognition performance, which allows us to exploit the two-part structure of the corresponding objective functions. In particular, we take advantage of the separability of the sparsity-inducing penalty functions to decompose the minimization problem into sub-problems that can be solved sequentially. Applying our method to a variety of state-of-the-art CNN models, our proposed method is able to simplify the original model, generating models with less computation and fewer parameters, while maintaining and often improving generalization performance. Accomplishments on a variety of models strongly verify that our proposed ADMM-based method can be a very useful tool for simplifying and improving deep CNNs. 
\end{abstract}

\section{Introduction}

Deep Convolutional Neural Networks (CNNs) have achieved remarkable performance in challenging computer vision problems such as image classification and object detection tasks, at the cost of a large number of parameters and computational complexity. These costs can be problematic for deployment especially on mobile devices and when real-time operation is needed. 

To improve the efficiency of CNNs, several attempts have been made to reduce the redundancy in the network. 
\cite{jaderberg2014speeding} proposed to represent the full-rank original convolutional filters tensor by a low-rank approximation composed of a sequence of two regular convolutional layers, with rectangular filters in the spatial domain. 
A different network connection structure is suggested by \cite{ioannou2015training}, which implicitly learns linear combinations of rectangular filters in the spatial domain, with different vertical/horizontal orientations.
\cite{tai2015convolutional} presented an exact and closed-form solution to the low-rank decomposition approach of \cite{jaderberg2014speeding} to enforce connection sparsity on CNNs. 

Sparse learning has been shown to be efficient at pruning the irrelevant parameters in many practical applications, by incorporating sparsity-promoting penalty functions into the original problem, where the added sparsity-promoting terms penalize the number of parameters (\cite{kiaee2016double,kiaee2016relevance,kiaee2016sparse}). Motivated by learning efficient architectures of a deep CNN for embedded implementations, our work focuses on the design of a sparse network using an initial pre-trained dense CNN. 

The alternating direction method of multipliers (ADMM) (\cite{boyd2011distributed}) has been extensively studied to minimize the augmented Lagrangian function for optimization problems, by breaking them into smaller pieces. It turns out that ADMM has been recently applied in a variety of contexts (\citet{lin2013design,shen2012distributed,meshi2011alternating}). We demonstrate that the ADMM provides an effective tool for optimal sparsity imposing on deep neural connections. This is achieved by augmenting a sparsity-inducing penalty term to the recognition loss of a pre-trained network. Different functions including the $l_0$-norm and its convex $l_1$-norm relaxations can be considered as a penalty term. The variables are then partitioned into two subsets, playing two different roles: 1) \textbf{promoting the sparsity} of the network at the level of a predetermined \textit{sparse block} structure; 2) \textbf{minimizing the recognition error}. 

The augmented Lagrangian function is then minimized with respect to each subset by fixing all other subsets at each iteration. In the absence of the penalty term, the performance results correspond to the original network with a dense structure. By gradually increasing the regularization factor of the sparsity-promoting penalty term, the optimal parameters move from their initial setting to the sparse structure of interest. This regularization factor is increased until the desired balance between performance and sparsity is achieved. 

Several approaches have been developed to create sparse networks by applying pruning or sparsity regularizers: \cite{wen2016learning,alvarez2016learning,liu2015sparse,han2015learning}. The most relevant to our work in these papers is the Structured Sparsity Learning (SSL) method of \cite{wen2016learning}, that regularizes the structures (i.e., filters, channels, filter shapes, and layer depth) of CNNs using a group lasso penalty function. However, the SSL approach suffers from two limitations compared to our proposed method. First, it relies on a rigid framework that disallows incorporation of non-differentiable penalty functions (e.g., $l_0$-norm). Second, it requires training the original full model, while our proposed method allows to decompose the corresponding optimization problems into two sub-problems and exploit the separability of the sparsity-promoting penalty functions to find an analytical solution for one of the sub-problems (see Sec.~\ref{section:conclusion} for more details).

Our numerical experiments on three benchmark datasets, namely CIFAR-10, CIFAR-100, and SVHN, show that the structure of the baseline networks can be significantly sparsified. 
While most previous efforts report a small drop or no change in performance, we found a slight increase of classification accuracy in some cases.  


\section{CNN with Sparse Filters}
\label{section:CNNS}

\begin{figure*}[t]
  \includegraphics[width=0.9\textwidth]{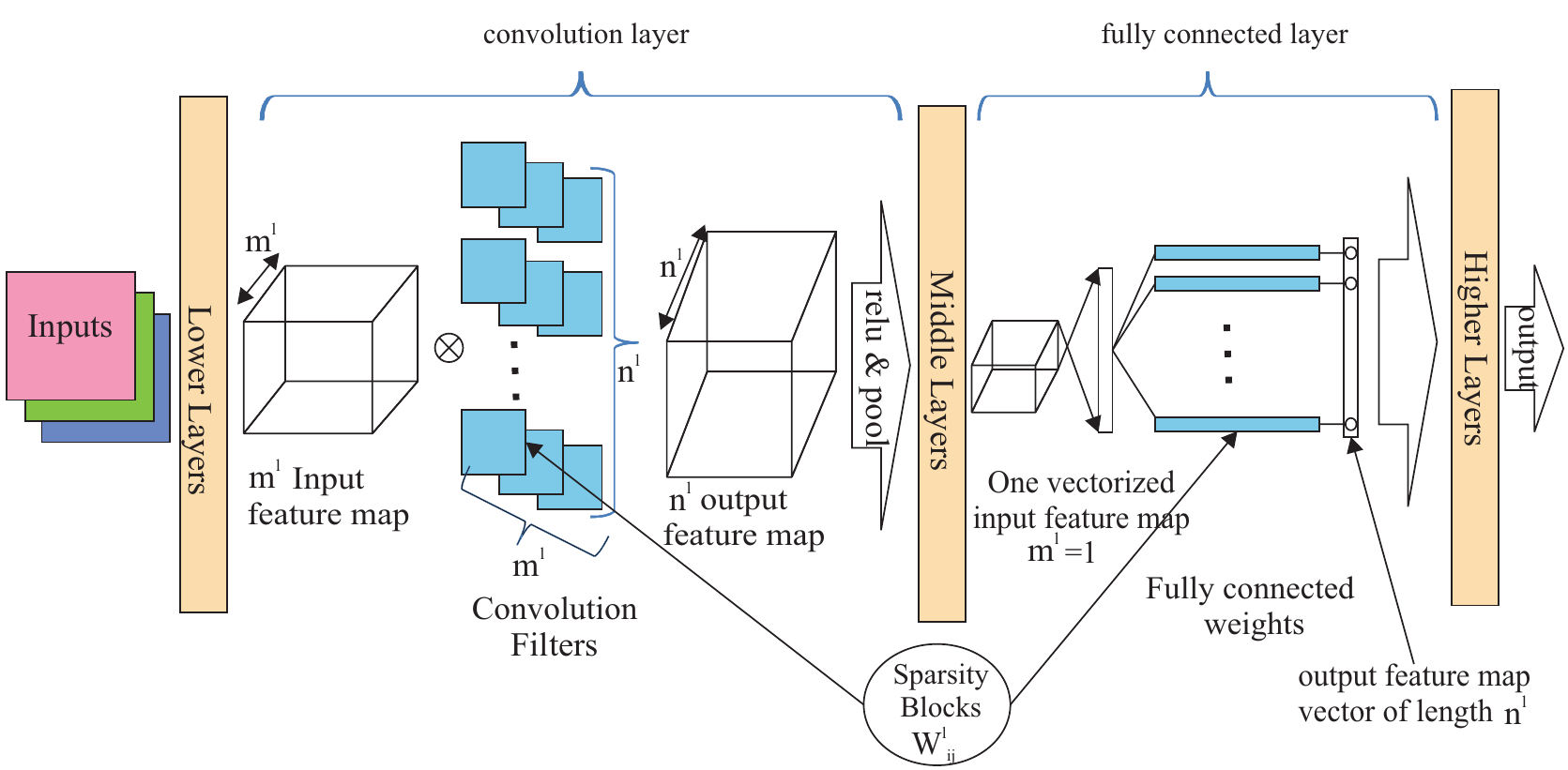}
  \centering
\caption{Architecture of a typical CNN, selected sparsity blocks at convolutional and fully connected layers are shown in blue.}
  \label{fig:SCNN}
\end{figure*}

Consider a CNN network consisting of a total of $L$ layers, including convolutional and fully connected layers, which are typically interlaced with rectified linear units and pooling (see Fig.~\ref{fig:SCNN}). 
Let the $l$-th layer includes $m^l$ input feature maps and $n^l$ output feature maps, with $\boldsymbol{\KernelLetter}^l_{ij}$ representing the convolution filter between the $i$-th and $j$-th input and output feature maps, respectively\footnote{Each fully connected layer can also be thought to be composed of several 1-dim convolutions, where the filter is of the same size as the input, hence is applied only at one location. In this context, if you look at the fully connected layer at Fig.~\ref{fig:SCNN}, you can see that it is just composed of one ($m^l=1$) vectorized input feature map and $n^l$ 1-dim convolutions, one for each output class.}. Our goal is to design the optimal filters, subject to sparse structural constraints. In order to obtain the filters which balance a trade-off between the minimization of the loss function and sparseness, we consider the following objective function
\begin{equation}
\label{eq:SP-objective}
\operatorname*{minimize}_{\boldsymbol{\KernelLetter}} ~\mathcal{L}_{net}(\boldsymbol{\KernelLetter}) + \mu f(\boldsymbol{\KernelLetter}),
\end{equation}
where $\mathcal{L}_{net}$ stands for the logistic loss function of the output layer of the network which is a function of the convolutional filters of all layers $\boldsymbol{\KernelLetter}=\{\boldsymbol{\KernelLetter}^l_{ij}|i=1,\ldots,m^l,j=1,\ldots,n^l,l=1,\ldots,L\}$. The term $f(\boldsymbol{\KernelLetter})$ is a penalty function on the total size of the filters. The $l_0$-norm (cardinality) function or relaxations to higher orders such as $l_1$-norm function can be employed to promote the sparsity of the filters.

The parameter $\mu$ controls the effect of sparse penalty term. As $\mu$ varies, the solution of (\ref{eq:SP-objective}) traces the trade-off path between the performance and the sparsity.
In the next section, the alternating direction method of multipliers (ADMM) which is employed to find the optimal solution of (\ref{eq:SP-objective}) is described.

\section{Using ADMM for Sparsifying CNNs}
\label{section:ADMM}
 
Consider the following constrained optimization problem:
\begin{align}
\label{eq:ADMM}
\operatorname*{minimize}_{\boldsymbol{\KernelLetter},\boldsymbol{F}} &\quad\mathcal{L}_{net}(\boldsymbol{\KernelLetter}) + \mu f(\boldsymbol{F}),  \nonumber  \\
\text{s.t.}&\quad\boldsymbol{\KernelLetter}-\boldsymbol{F}=\boldsymbol{0},   
\end{align}
which is clearly equivalent to the problem stated in (\ref{eq:SP-objective}). The key point here is that by introducing an additional variable $\boldsymbol{F}$ and an additional constraint $\boldsymbol{\KernelLetter}-\boldsymbol{F}=\boldsymbol{0}$, the objective function of the problem (\ref{eq:SP-objective}) is decoupled into two parts that depend on two different variables. 

The augmented Lagrangian associated with the constrained problem (\ref{eq:ADMM}) is given by
\begin{align}
\label{eq:augmented}
\mathcal{C}(\boldsymbol{\KernelLetter},&\boldsymbol{F},\boldsymbol{\Gamma}) =\mathcal{L}_{net}(\boldsymbol{\KernelLetter}) + \mu f(\boldsymbol{F}) \nonumber \\
&+\sum_{l,i,j}\text{trace}({\boldsymbol{\Gamma}^l_{ij}}^{T}(\boldsymbol{\KernelLetter}^l_{ij}-\boldsymbol{F}^l_{ij})) +\frac{\rho}{2}\sum_{l,i,j}\parallel \boldsymbol{\KernelLetter}^l_{ij}-\boldsymbol{F}^l_{ij}\parallel^{2}_{F},
\end{align}
where $\boldsymbol{\Gamma}^l_{ij}$ is the dual variable (i.e., the Lagrange multiplier), $\rho$ is a positive scalar, $\parallel.\parallel_{F}$ and is the Frobenius norm. 

In order to find a minimizer of the constrained problem (\ref{eq:augmented}), the ADMM algorithm uses a sequence of iterative computations:
\begin{enumerate}
\item Make use of a descent method to solve the following \textbf{performance promoting} problem,
\begin{equation}
\label{eq:PP}
{\boldsymbol{\KernelLetter}}^{\{k+1\}}=\operatorname*{arg\,min}_{\boldsymbol{\KernelLetter}} \mathcal{C}\left(\boldsymbol{\KernelLetter},{\boldsymbol{F}}^{\{k\}},{\boldsymbol{\Gamma}}^{\{k\}}\right);
\end{equation}

\item Find the analytical expressions for the solutions of the following \textbf{sparsity promoting} problem,
\begin{equation}
\label{eq:SP}
{\boldsymbol{F}}^{\{k+1\}} =\operatorname*{arg\,min}_{\boldsymbol{F}}\mathcal{C}\left({\boldsymbol{\KernelLetter}}^{\{k+1\}},\boldsymbol{F},{\boldsymbol{\Gamma}}^{\{k\}}\right );
\end{equation}
\item  Update the dual variable $\boldsymbol{\Gamma}^l_{ij}$ using a step-size equal to $\rho$, in order to guarantee that the dual feasibility conditions is satisfied in each ADMM iteration,
\begin{align}
\label{eq:dual}
{\boldsymbol{\Gamma}^l_{ij}}^{\{k+1\}} ={\boldsymbol{\Gamma}^l_{ij}}^{\{k\}} +\rho \left ({\boldsymbol{\KernelLetter}^l_{ij}}^{\{k+1\}}-{\boldsymbol{F}^l_{ij}}^{\{k+1\}}\right).
\end{align}
\end{enumerate}

The three described computation steps are applied in an alternating manner. Re-estimation stops when the Frobenius distance of $\boldsymbol{F}$ in two consecutive iterations as well as the Frobenius distance of $\boldsymbol{\KernelLetter}$ and $\boldsymbol{F}$ at current iterations are less than a small threshold value. The details of steps 1 and 2 are described in the next sections. The outline of the proposed sparse CNN approach is summarized in Algorithm \ref{alg:SCNN}. At each individual regularization $\mu$, in order to improve the performance of the sparse-structured network we fine tune the initial non-augmented recognition loss subject to the parameters belonging to the identified sparse structure. 
\begin{algorithm}[t]
\caption{Outline of the proposed sparse CNN algorithm}
\label{alg:SCNN}
\begin{algorithmic}[1]
\Function{sparse-CNN}{data, model}
\State Set {$\boldsymbol{\KernelLetter}$ to a pre-trained reference CNN model 
\State $\boldsymbol{\Gamma} = \boldsymbol{0}$, $ \boldsymbol{F}=  \boldsymbol{\KernelLetter}$}
\State $\mathcal{S}$: a set of small logarithmically spaced points in increasing order, as regularization factor.
\For{\textbf{each} $\mu$ in $\mathcal{S}$}
\Do
\State Find the estimate of ${\boldsymbol{\KernelLetter}}^{\{k+1\}}$ by minimizing (\ref{eq:PP_equivalent})
\State Find the estimate of ${\boldsymbol{F}}^{\{k+1\}}$ from (\ref{eq:norm1}) or (\ref{eq:norm0})
\State Update dual variable ${\boldsymbol{\Gamma}}^{\{k+1\}}$ from (\ref{eq:dual})
\doWhile{$\parallel {\boldsymbol{\KernelLetter}}^{\{k+1\}} - {\boldsymbol{F}}^{\{k+1\}}\parallel_{F} > \epsilon$ or $ \parallel {\boldsymbol{F}}^{\{k+1\}} - {\boldsymbol{F}}^{\{k\}} \parallel_{F} > \epsilon$}
\State Fix the identified sparse structure and fine-tune network according to $\mathcal{L}_{net}$ w.r.t. non-zero parameters
\EndFor\\
\Return $\boldsymbol{\KernelLetter}^l_{ij}$
\EndFunction
\end{algorithmic}
\end{algorithm}

\subsection{Performance Promoting Step}
\label{section:PP}
By completing the squares with respect to $\boldsymbol{\KernelLetter}$ in the augmented Lagrangian $\mathcal{C}(\boldsymbol{\KernelLetter},\boldsymbol{F},\boldsymbol{\Gamma})$, we obtain the following equivalent problem to (\ref{eq:PP}) 
\begin{align}
\label{eq:PP_equivalent}
\operatorname*{minimize}_{\boldsymbol{\KernelLetter}}~\mathcal{L}_{net}(\boldsymbol{\KernelLetter}) +\frac{\rho}{2}\sum_{l,i,j}\parallel \boldsymbol{\KernelLetter}^l_{ij}-\boldsymbol{U}^l_{ij}\parallel^2_{F},
\end{align}
where $ \boldsymbol{U}^l_{ij}= \boldsymbol{F}^l_{ij}-\frac{1}{\rho}\boldsymbol{\Gamma}^l_{ij}$. From (\ref{eq:PP_equivalent}), it can be seen that by exploiting the separability property of ADMM method in the minimization of the augmented Lagrangian, the sparsity penalty term which might be non-differentiable is excluded from (\ref{eq:PP_equivalent}). Consequently, descent algorithms that rely on the differentiability can be utilized to solve the performance promoting sub-problem (\ref{eq:PP_equivalent}) 

This property allows that popular software and toolkit resources for Deep Learning, including Caffe, Theano, Torch, and TensorFlow, to be employed for implementing the proposed approach. In our work, we use Stochastic Gradient Descent (SGD) method of TensorFlow to optimize the weights ($\boldsymbol{\KernelLetter}$), which seemed a reasonable choice for the high-dimensional optimization problem at hand. The entire procedure relies mainly on the standard forward-backward pass that is used to train the convolutional network.

\subsection{Sparsity Promoting Step}
\label{section:SP}

The completion of squares with respect to $\boldsymbol{F}$ in the augmented Lagrangian can be used to show that (\ref{eq:SP}) is equivalent to 
\begin{align}
\label{eq:SP_equivalent}
\operatorname*{minimize}_{\boldsymbol{F}}\mu f(\boldsymbol{F})+\frac{\rho}{2}\sum_{l,i,j}\parallel \boldsymbol{F}^l_{ij}-\boldsymbol{V}^l_{ij}\parallel^2_{F},
\end{align}
where $\boldsymbol{V}^l_{ij}=\boldsymbol{\KernelLetter}^l_{ij}+\frac{1}{\rho}\boldsymbol{\Gamma}^l_{ij}$. From (\ref{eq:SP_equivalent}), it can be seen that the proposed method provides a flexible framework to select arbitrary \textit{sparsity blocks}. Sparse structure can then be achieved at the level of the selected block. Specifically, both terms on the right-hand side of (\ref{eq:SP_equivalent}), $f(\boldsymbol{F})$ (for either the case of $l_1$-norm or $l_0$-norm) as well as the square of the Frobenius norm can be written as a summation of component-wise functions of a tensor. In our experiments, individual filter components are selected as the sparsity blocks (see Fig.~\ref{fig:SCNN}). Hence (\ref{eq:SP_equivalent}) can simply be expressed in terms of $\boldsymbol{F}^l_{ij}$ components corresponding to the filters. However, any other individual sub-tensor components can be selected as the sparsity block.

More precisely, if $f(\boldsymbol{F})$ is selected to be the $l_1$-norm function, then $\mathcal{C}(\boldsymbol{F}) =\sum_{l,i,j}(\mu\parallel\boldsymbol{F}^l_{ij}\parallel_{F} +\frac{\rho}{2}\parallel \boldsymbol{F}^l_{ij}-\boldsymbol{V}^l_{ij}\parallel^2_{F})$ and consequently (\ref{eq:SP_equivalent}) is converted to a minimization problem that only involves spatial filters. The solution of  (\ref{eq:SP_equivalent}) can then be determined
analytically by the following soft thresholding operation,
\begin{align}
\label{eq:norm1}
 {\boldsymbol{F}^l_{ij}}^{*}= \left\{ 
  \begin{array}{l l}
   \left(1-\frac{a}{\parallel\boldsymbol{V}^l_{ij}\parallel_{F}}\right)\boldsymbol{V}^l_{ij},  &\quad \text{if~}\parallel\boldsymbol{V}^l_{ij}\parallel_{F}>a\\
   0, & \quad\text{otherwise}\\
  \end{array}\right.,
\end{align}
where $a = \frac{\mu}{\rho}$. Similarly, the following hard thresholding operation is the analytical solution for the case of the selection of the $l_0$-norm $f(\boldsymbol{F})$ penalty term.
\begin{align}
\label{eq:norm0}
 {\boldsymbol{F}^l_{ij}}^{*}= \left\{ 
  \begin{array}{l l}
   \boldsymbol{V}^l_{ij},  & \quad\text{if~}\parallel\boldsymbol{V}^l_{ij}\parallel_{F}>b\\
   0, & \quad\text{otherwise}\\
  \end{array}\right.,
\end{align}
where $b = \sqrt{\frac{2\mu}{\rho}}$.

\subsection{Convergence of the Proposed ADMM-based Sparse CNN Method} 
For convex problems, the ADMM is guaranteed to converge to the global optimum solution (\cite{boyd2011distributed}). For non-convex problems, where there is a general lack of theoretical proof, extensive computational experience suggests that ADMM works well when the penalty parameter $\rho$ in the augmented Lagrangian is chosen to be sufficiently large. This is related to the quadratic term that tends to locally convexify the objective function for sufficiently large $\rho$. 

Unfortunately, in the deep learning problems, the objective is inherently highly non-convex and consequently there is the risk that it becomes trapped into a local optimum. This difficulty could be circumvented by considering a warm start that may be obtained by running a pre-trained version of the network. The proposed ADMM approach is then used to sparsify the final solution.
Using this procedure, as the experiments in the next section show, we have obtained good empirical results. 

\section{Experimental Results} 
\label{section:experiments}
In order to validate our approach, we show that our proposed sparse CNN approach can be efficiently applied to existing state-of-the-art network architectures to reduce the computational complexity without reducing the accuracy performance. For this purpose, we evaluate the proposed scheme on the CIFAR-10, CIFAR-100, and SVHN datasets with several CNN models. 

In the implementation of the performance promoting step in Sec.~\ref{section:PP}, the batch size is $128$ and the learning rate is set to a rather small value (i.e., $0.001$ to search the space around the dense initialized filters to find a sparse solution). Since the regularization factor $\mu$ is selected from gradually increasing values, for the first small values of $\mu$ the selection of long epochs for performance-promoting step (inner loop) and fine-tuning steps is computationally prohibitive and would result in over-fitting. Instead, we start with one epoch for the first $\mu$ and increase the number of epochs by $\delta$ for the next $\mu$ values up to the $\nu$-th $\mu$ value, after which the number of epochs is limited to $\delta\nu$. We found that $\delta=1$ and $\nu=15$ generally work well in our experiments. We already incorporated the number of training epochs at tables \ref{tab:CIFAR10}, \ref{tab:CIFAR100}, and \ref{tab:SVHN} of Appendix \ref{sec:AppendixB}. If the maximum limiting number of iterations of inner loop is $\xi$ (suggested value of $\xi$=10), the training time of the $\nu$-th $\mu$ value takes a total of $\delta\nu\xi+\delta\nu$ epochs ($\delta\nu\xi$ for performance-promoting step and $\delta\nu$ for fine-tuning) under the worst-case assumption, where the inner loop has not converged and completes only at the $\xi$-th iteration.

\subsection{Results on CIFAR-10 Object Classification}
\begin{figure*}[t]
  \includegraphics[width=0.95\textwidth]{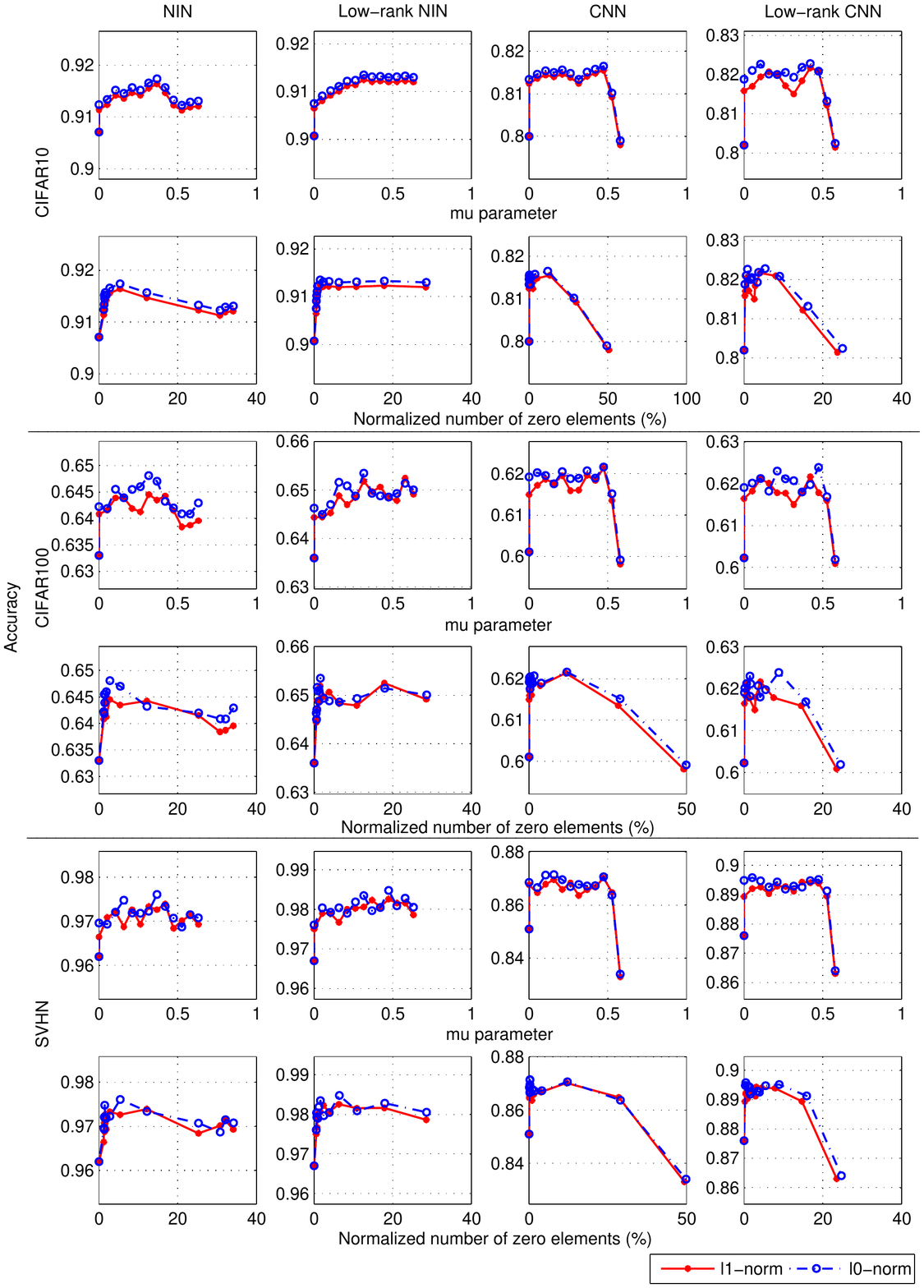}
  \centering
\caption{Variation of accuracy measure against: (odd rows) values of $\mu_0$ parameters and (even rows) normalized number of zero elements for different models and datasets.}
  \label{fig:sparsity_plot}
\end{figure*}

The CIFAR-10 dataset is a well-known small dataset of 60,000 32 x 32 images in 10 classes. This dataset comprises standard sets of 50,000 training images, and 10,000 test images. As a baseline for the CIFAR-10 dataset, we deploy four models: the Network in Network (NIN) architecture \citep{lin2013network}, its low-rank version \citep{ioannou2015training}, a custom CNN, and its low-rank counterpart as well, two last being learned from scratch on the CIFAR dataset. The configurations of the baseline models are outlined in Table \ref{tab:config}.
\begin{table}[!ht]
\caption{Structure of the baseline networks.}
\label{tab:config}

\begin{subtable}{0.55 \textwidth}
\begin{tabular}{|c|c|c|}
\hline
 & NIN  & Low-rank NIN   \\
\hline
\multirow{2}{*}{conv1} & \multirow{2}{*}{$ 3 \times 3 \times 192$} & h: $1 \times 3 \times 96$ \\
& &v: $3 \times 1 \times 96$\\
\hline
conv2,3  &\multicolumn{2}{c|}{$1 \times 1 \times 160$, $1 \times 1 \times 96$  }\\
\hline
\multirow{2}{*}{conv4} & \multirow{2}{*}{$3 \times 3 \times 192$} & h: $1 \times 3 \times 96$ \\
  & &v: $3 \times 1 \times 96$\\
\hline
\multirow{2}{*}{conv5} & \multirow{2}{*}{$3 \times 3 \times 192$} & h: $1 \times 3 \times 96$ \\
  & &v: $3 \times 1 \times 96$\\\hline
conv6,7  &\multicolumn{2}{c|}{$1 \times 1 \times 192$, $1 \times 1 \times 192$} \\
\hline
\multirow{2}{*}{conv8} & \multirow{2}{*}{$3 \times 3 \times 192$} & h: $1 \times 3 \times 96$ \\
& &v: $3 \times 1 \times 96$\\
\hline
conv9,10 &\multicolumn{2}{c|}{$1 \times 1 \times 192$, $1 \times 1 \times 10$} \\
\hline
\end{tabular}

\end{subtable}%
\begin{subtable}[b]{0.55\textwidth}
\begin{tabular}{|c|c|c|}
\hline
 & CNN  & Low-rank CNN   \\
\hline
\multirow{2}{*}{conv1} & \multirow{2}{*}{$3\times3\times 96$}  &h: $1 \times 3\times 48$ \\
& &v: $3 \times 1 \times 46$\\
\hline
\multirow{2}{*}{conv2} & \multirow{2}{*}{$3 \times 3 \times 128$}  &h: $1 \times 3 \times 64$ \\
& &v: $3 \times 1 \times 64$\\
\hline
\multirow{2}{*}{conv3} & \multirow{2}{*}{$3 \times 3 \times 256$}  &h: $1 \times 3 \times128$ \\
& &v: $3 \times 1 \times 128$\\
\hline
\multirow{2}{*}{conv4} & \multirow{2}{*}{$3 \times 3 \times 64$}  &h: $1 \times 3 \times 32$ \\
& &v: $3 \times 1 \times 32$\\
\hline
fc1  &$1024\times 256$ &$1024\times 256$  \\
\hline
fc2  &$256 \times10$ &$256 \times10$\\
\hline
\end{tabular}
\end{subtable}%
\end{table}
The architecture of the NIN model is slightly different from the one introduced in \cite{lin2013network}. The original NIN uses 5x5 filters in the first and second convolutional layer which are replaced with one and two layers of 3x3 filters, respectively. As suggested by \cite{ioannou2015training}, this modified architecture has comparable accuracy and less computational complexity. In the low-rank networks, every single convolutional layer of the full-rank model is replaced with two convolutional layers with horizontal and vertical filters. NIN and low-rank NIN have an accuracy of $90.71\,\%$ and $90.07\,\%$, respectively. The custom CNN and its low-rank variant show a baseline accuracy of $80.0\,\%$ and $80.2\%$, respectively. The results of our experiments are plotted in Fig.~\ref{fig:sparsity_plot} for both $l_0$-norm and $l_1$-norm sparsity constraints.   

Fig.~\ref{fig:sparsity_plot} shows how the accuracy performance changes as we increase the regularization factor $\mu$. The case with $\mu=0$ can be considered as the baseline model. In order to avoid over pruning of some layers, if the number of pruned filters in one layer exceeds $50\,\%$ of the total number of filters in that layer, then we change the pruning threshold to the statistical mean of the Frobenius norm of all the filters at that layer in the sparsity promoting step (explained in Sec.~\ref{section:SP}) to stop the over pruning of that layer.
Taking the NIN and low-rank-NIN as an example, using the $l_0$-norm sparsity function, the parameters in the networks are reduced by $34.13\,\%$ and $28.5\,\%$ and the relative accuracy performance is $+.5\,\%$ and $+1.23\,\%$, respectively. Using the $l_1$-norm sparsity constraint achieves slightly lower accuracy compared to the $l_0$-norm, although it still conveniently sparsifies the network. 

Using the proposed sparsity promoting approach on the custom CNN models, the networks with sparse connections and similar accuracy ($79.9\%$ vs $80\,\%$) are achieved, but they have approximately $49.4\,\%$ fewer parameters than the original networks model. Since the target solution is likely to be sparse, enforcing sparsity at the beginning of the learning process with our proposed method provides a way to avoid overfitting for achieving a better performance. However, as the experiment results show, increasing more the sparsity strength of the solution may lead to slight oversmoothing and drop in the performance. For the low-rank CNN, we achieve a comparable accuracy of $80.14\,\%$, with $25\,\%$ fewer parameters. 

To further verify that the advantage of ADMM training is statistically significant, a $t$-test is conducted by repeating the experiment 15 times on CIFAR-10 by using NIN model. The $t$-test results are in Appendix \ref{sec:AppendixA}. In Appendix \ref{sec:AppendixB}, however, we present detailed results for random sample runs over the configurations tested. According to the results presenting in Table~\ref{tab:CIFAR10} of Appendix \ref{sec:AppendixB}, the number of parameters in the network can be reduced by a large factor, especially for the higher convolution layers. Interestingly, even with significant reductions in the number of parameters, the performance does not decrease that much. This parameter reduction also gives rise to the speed-up of the network, reported at the last columns of the tables. Note that most of the results listed in Table~\ref{tab:CIFAR10} outperform the baseline model. 

\subsection{Results on CIFAR-100 Object Classification} 

The CIFAR-100 dataset is similar to the CIFAR-10 dataset containing 100 classes with 600 images per class. For CIFAR-100 we again use the baseline networks in Table~\ref{tab:config} with only one structural difference (i.e., the NIN networks contain 100 feature maps at the last convolution layer and custom CNN networks contain 100 output labels). 
The baseline NIN, low-rank NIN, custom CNN, and low-rank CNN models show a test accuracy of $63.3\,\%$, $63.6\,\%$, $60.11\,\%$, and $60.23\,\%$, respectively. Using the proposed sparsity promoting approach on these networks, the total number of parameters in the layers can be reduced by a large factor with comparable or even better performance accuracy.  

In particular, on the CIFAR-100 dataset, we achieve $64.09\,\%$ classification accuracy with $34.1\,\%$ sparsity for the NIN model, which improves upon the original NIN on this dataset. A test accuracy of $65.23\,\%$ is obtained for CIFAR-100 for the low-rank NIN model with $28.5\,\%$ sparsity which surpasses the performance of the baseline model. The proposed method on custom CNN and low-rank CNN show comparable performance accuracy to their corresponding baseline models ($59.82\,\%$ vs  $60.11\,\%$ and $60.1\,\%$ vs  $60.23\,\%$) with much less computation ($49.7\,\%$ and $24.4\,\%$ number of zero elements, respectively).
The details of changing sparsity in different layers of the networks on the CIFAR-100 dataset are presented in Table~\ref{tab:CIFAR100} of Appendix \ref{sec:AppendixB}. The same conclusions made for CIFAR-10 can be drawn from these results.

\subsection{Results on SVHN Object Classification} 

The SVHN dataset consists of 630,420 32x32 color images of house numbers collected by Google Street View. The task of this dataset is to classify the digit located at the center of each image. The structure of the baseline models used in SVHN is similar to those used for CIFAR-10, which are presented in Table \ref{tab:config}. The training and testing procedure of the baseline models follows \cite{lin2013network}. The baseline NIN, low-rank NIN, custom CNN, and low-rank CNN models show the accuracy of $96.2\,\%$, $96.7\,\%$, $85.1\,\%$, and $87.6\,\%$, respectively. For this dataset, by applying our proposed sparse approach to NIN and low-rank NIN models, we obtain a higher accuracy of $96.97\,\%$ and $99\,\%$  with $34.17\,\%$ and $28.6\,\%$ fewer parameters, respectively. We also achieve comparable accuracy of $83.3\,\%$  and $86.3\,\%$ using $49.7\,\%$ and $24.7\,\%$ less parameters of the original model parameters on custom CNN and low-rank CNN models, respectively (see Table~\ref{tab:SVHN} of Appendix \ref{sec:AppendixB} for the details on changing the sparsity in different layers of the networks on SVHN dataset).

\section{Discussion} 
\label{section:conclusion}

In this paper we proposed a framework to optimal sparsification of a pre-trained CNN approach. We employed the ADMM algorithm to solve the optimal sparsity-promoting problem, whose solution gradually moves from the original dense network to the sparse structure of interest as our emphasis on the sparsity-promoting penalty term is increased. The proposed method could potentially reduce the memory and computational complexity of the CNNs significantly.

Briefly, the main contributions of the proposed sparse CNN can be summarized as follows:
\begin{description}
\item[Separability]: The penalty function is separable with respect to the individual elements of the weight tensors. In contrast, the recognition loss function cannot be decomposed into component-wise functions of the weight tensors. By separating the two parts in the minimization of the augmented Lagrangian, we can analytically determine the solution to the sparsity promoting sub-problem (\ref{eq:SP_equivalent}).
\item[Differentiability]: The recognition loss function $\mathcal{L}_{net}(\boldsymbol{\KernelLetter})$ is typically differentiable with respect to the parameters, as opposed to some choices of sparsity penalty terms (e.g., $l_0$-norm which is a non-differentiable function). In our approach, by separating the two parts in the minimization of the augmented Lagrangian, descent algorithms can be utilized to solve the performance promoting sub-problem (\ref{eq:PP_equivalent}) while different functions (e.g., $l_0$-norm and $l_1$-norm) can be incorporated as means of sparsity penalty terms in the original problem (\ref{eq:SP-objective}).   
\item[Model size reduction]: There are recent works focusing on reducing the parameters in the convolutional layers (\cite{jaderberg2014speeding,ioannou2015training,tai2015convolutional}). In CNN models, the model size is dominated by the fully connected layers. Thus, the previous approaches are not capable of reducing the size of the whole model. Our proposed approach can be applied on both the convolution and fully connected layers and can speed up the computation as well as compressing the size of the model.
\item[Combinability with other methods]: Several attempts have been made to compress the deep networks using the weights sharing and quantization (\cite{han2016deep,gupta2015deep,vanhoucke2011improving}). However, these techniques can be used in conjunction with our proposed sparse method to achieve further speedup. 
\end{description}

Some methods such as SSL (\cite{wen2016learning}), based on group Lasso regularization of the block structures (e.g. filters), appears to be closely related to our work. Indeed, these methods learn sparse filters and minimize the classification error simultaneously. In contrast, our proposed approach uses ADMM to provide a separate scheme that optimize the sparse blocks and classification error separately. Indeed, at the core of our contribution, ADMM brings the above major \textit{separability} and \textit{differentiability} benefits to the proposed sparse CNN method. Our proposed algorithm has the advantage that it is partially and analytically solvable due to the \textit{separability} property. This contributes to the efficient trainability of the model. Moreover, the \textit{differentiability} problem of $l_0$-norm penalty function makes it unusable for a joint performance/sparsity optimization, while it can be conveniently incorporated as a sparsity penalty term in our proposed method.

Furthermore, in the SSL method, strengths of structured sparsity regularization is selected by cross validation and the networks weights are initialized by the baseline. This is computationally beneficial for small regularization level. However, for larger regularization value, the presented SSL approach requires training the original full model from scratch. In contrast, our approach gradually modifies the regularization factor and each step continues training from the solution achieved in the previous step (corresponding to the previous regularization factor), which plays an important role in reducing the computational complexity of the method.


\section*{Acknowledgments}
The authors gratefully acknowledge financial support by NSERC-Canada, MITACS, and E Machine Learning Inc., a GPU grant from NVidia, and access to the computational resources of Calcul Qu\'ebec and Compute Canada. The authors are also grateful to Annette Schwerdtfeger for proofreading this manuscript.

\bibliography{iclr2017_conference}
\bibliographystyle{iclr2017_conference}

\begin{appendices}

\section{Significance Validation of Improvements}
\label{sec:AppendixA}

On order to verify that the advantage of ADMM training is statistically significant, we conduct $t$-test by repeating the experiment 15 times on CIFAR-10 using NIN to compare the error rate of ADMM training and standard fine-tuning (by dropping the learning rate upon ``convergence'' and continuing to learn), with the same number of epochs and learning rates. Initialized from the same baseline model with $90.71\,\%$ accuracy, the ADMM training using $l_0$-norm and standard fine-tuning on average achieve accuracy of $91.34\,\%$ and $91.09\,\%$, respectively. The results demonstrate the ADMM training achieves improvement of $0.63\,\%$ from the baseline model which is statistically significant ($t$-test result with $p<0.001$). ADMM training performance is also significantly $25\,\%$ better than what the standard fine-tuning achieves ($t$-test result with $p<0.001$). The $t$-test experiment also shows that ADMM could reduce the variance of learning. In the 15 repeated experiments, ADMM training has the lowest standard deviation of errors compared with their counterparts using standard fine-tuning (standard deviation of $0.04\,\%$ for ADMM vs $0.06\,\%$ for standard fine-tuning).

\begin{table}[h]
\centering
\caption{$t$-test results for the significance validation of the performances. Results are reported over 15 runs on CIFAR-10 using NIN.}
\begin{tabular}{|c|c|c|c|}
\hline
 ~ & ADMM training & Standard fine-tuning  & Baseline \\
\hline
Mean accuracy ($\%$)  &91.34 &91.09 &90.71\\\hline
Accuracy standard deviation ($\%$)  &0.04 &0.06 &--\\\hline
Sparsity ($\%$)  &34.5 &0 &0\\\hline
$p$-value  &-- &0.001 &0.001\\
\hline
\end{tabular}
\label{tab:datasets}
\end{table}

\section{Single Run Results}
\label{sec:AppendixB}

Due to space consideration, we present some extra results in the current appendix. First, the results for our sparsity promoting approach for the different models on CIFAR-10, CIFAR-100, and SVHN are presented in Tables \ref{tab:CIFAR10}, \ref{tab:CIFAR100} and \ref{tab:SVHN}, respectively. Follows in Table \ref{tab:ResultsCombined} results showing joint variations of accuracy and sparsity obtained with increasing $\mu$ values, for the three tested datasets. All these results are for a single random run of each method on the dataset at hand.

\begin{table}[!htbp]
\centering
\caption{Performance of the proposed ADMM-based sparse method on the CIFAR-10 dataset.}
\label{tab:CIFAR10}
\begin{subtable}{0.75\textwidth}
\caption{NIN model}
\resizebox{\columnwidth}{!}{%
  \begin{threeparttable}
\begin{tabular}{|c|c|c|c|c|c|c|}
\hline
~&$\mu$ &Accuracy ($\%$)&Filter ($\#$)\tnote{*}&Sparsity ($\%$)&Training epochs ($\#$)&Speedup\\ \hline 
 \multirow{8}{*}{\rotatebox{90}{ $l_1$-norm}}
&0&90.71 &0-0-0-0&0.00&0&1.00\\ \cline{2-7} 
&0.000&91.14 &33-482-968-0&1.23&4&1.06\\ \cline{2-7} 
&0.105&91.42 &33-551-1027-0&1.36&16&1.19\\ \cline{2-7} 
&0.211&91.47 &34-609-1144-31&1.55&30&1.31\\ \cline{2-7} 
&0.316&91.56 &41-749-1428-589&2.72&48&1.44\\ \cline{2-7} 
&0.421&91.47 &56-1822-2300-5630&12.14&90&1.56\\ \cline{2-7} 
&0.526&91.13 &70-6810-4834-12451&30.73&120&1.69\\ \cline{2-7} 
&0.632&91.21 &107-6810-11568-12451&34.07&140&1.81\\ \cline{2-7} 
\hline
 \multirow{8}{*}{\rotatebox{90}{ $l_0$-norm}}
&0&90.71 &0-0-0-0&0.00&0&1.00\\ \cline{2-7} 
&0.000&91.24 &34-482-969-2&1.23&4&1.06\\ \cline{2-7} 
&0.105&91.52 &36-554-1031-2&1.37&16&1.19\\ \cline{2-7} 
&0.211&91.57 &39-614-1148-35&1.57&36&1.31\\ \cline{2-7} 
&0.316&91.66 &46-755-1432-596&2.75&64&1.44\\ \cline{2-7} 
&0.421&91.57 &65-1828-2304-5640&12.18&80&1.56\\ \cline{2-7} 
&0.526&91.23 &81-6821-4843-12461&30.78&96&1.69\\ \cline{2-7} 
&0.632&91.31 &118-6821-11577-12461&34.12&112&1.81\\ \cline{2-7} 
\hline
\end{tabular}
  \begin{tablenotes}
  \item[*] In the order of first hidden layer to last hidden layer out of a total of 576-18432-36864-36864 filters, respectively.
  \end{tablenotes}
  \end{threeparttable}
  }
\end{subtable}

\vspace{1em}
\begin{subtable}{0.75\textwidth}
\caption{Low-rank NIN model}
\resizebox{\columnwidth}{!}{%
 \begin{threeparttable}
\begin{tabular}{|c|c|c|c|c|c|c|}
\hline
~&$\mu$ &Accuracy ($\%$)&Filter ($\#$)\tnote{*}&Sparsity ($\%$)&Training epochs ($\#$)&Speedup\\ \hline 
 \multirow{8}{*}{\rotatebox{90}{ $l_1$-norm}}
&0&90.07 &0-0-0-0-0-0-0-0&0.00&0&1.00\\ \cline{2-7} 
&0.000&90.65 &8-9-192-96-102-96-0-1&0.54&4&1.09\\ \cline{2-7} 
&0.105&90.92 &9-9-192-98-180-97-20-7&0.66&16&1.26\\ \cline{2-7} 
&0.211&91.12 &9-9-201-104-287-116-78-14&0.88&30&1.43\\ \cline{2-7} 
&0.316&91.25 &11-10-275-135-483-177-270-58&1.53&56&1.61\\ \cline{2-7} 
&0.421&91.22 &15-22-479-239-1105-411-983-225&3.75&100&1.78\\ \cline{2-7} 
&0.526&91.21 &19-28-1163-644-2832-1343-3083-871&10.76&120&1.96\\ \cline{2-7} 
&0.632&91.20 &30-37-2707-1989-6509-4176-7681-3232&28.43&140&2.13\\ \cline{2-7} 
\hline
 \multirow{8}{*}{\rotatebox{90}{ $l_0$-norm}}
&0&90.07 &0-0-0-0-0-0-0-0&0.00&0&1.00\\ \cline{2-7} 
&0.000&90.75 &9-10-194-96-103-98-2-2&0.55&4&1.09\\ \cline{2-7} 
&0.105&91.02 &10-10-194-102-182-99-23-8&0.68&16&1.26\\ \cline{2-7} 
&0.211&91.22 &13-11-204-110-293-119-81-15&0.91&36&1.43\\ \cline{2-7} 
&0.316&91.35 &18-16-281-141-490-181-277-59&1.58&48&1.61\\ \cline{2-7} 
&0.421&91.32 &23-30-485-245-1112-420-990-233&3.82&80&1.78\\ \cline{2-7} 
&0.526&91.31 &29-36-1173-651-2839-1354-3092-879&10.84&108&1.96\\ \cline{2-7} 
&0.632&91.30 &40-46-2719-1996-6519-4188-7692-3240&28.51&126&2.13\\ \cline{2-7} 
\hline
\end{tabular}
  \begin{tablenotes}
  \item[*] In the order of first hidden layer to last hidden layer out of a total of 288-288-9216-9216-18432-18432-18432-18432  filters, respectively.
  \end{tablenotes}
  \end{threeparttable}
 }
\end{subtable}

\vspace{1em}
\begin{subtable}{0.75\textwidth}
\caption{CNN model}
\resizebox{\columnwidth}{!}{%
 \begin{threeparttable}
\begin{tabular}{|c|c|c|c|c|c|c|}
\hline
~&$\mu$ &Accuracy ($\%$)&Filter ($\#$)\tnote{*}&Sparsity ($\%$)&Training epochs ($\#$)&Speedup\\ \hline 
 \multirow{8}{*}{\rotatebox{90}{ $l_1$-norm}}
&0&80.00 &0-0-0-0-0&0.00&0&1.00\\ \cline{2-7} 
&0.000&81.24 &0-0-0-0-0&0.00&4&1.00\\ \cline{2-7} 
&0.105&81.44 &0-0-0-0-0&0.00&16&1.00\\ \cline{2-7} 
&0.211&81.46 &0-0-0-0-0&0.00&30&1.00\\ \cline{2-7} 
&0.316&81.24 &0-5-20-9-57&2.40&64&1.21\\ \cline{2-7} 
&0.421&81.48 &0-267-843-792-57&5.11&80&1.37\\ \cline{2-7} 
&0.526&80.92 &3-2870-8922-7161-57&29.82&120&1.94\\ \cline{2-7} 
&0.579&79.80 &7-5383-17189-10736-57&50.63&130&2.15\\ \cline{2-7} 
\hline
 \multirow{8}{*}{\rotatebox{90}{ $l_0$-norm}}
&0&80.00 &0-0-0-0-0&0.00&0&1.00\\ \cline{2-7} 
&0.000&81.34 &1-2-2-1-1&0.05&4&1.01\\ \cline{2-7} 
&0.105&81.54 &1-2-3-2-3&0.14&16&1.02\\ \cline{2-7} 
&0.211&81.56 &5-3-5-5-5&0.23&36&1.03\\ \cline{2-7} 
&0.316&81.34 &5-8-25-17-21&0.95&56&1.10\\ \cline{2-7} 
&0.421&81.58 &10-273-853-800-23&3.75&80&1.30\\ \cline{2-7} 
&0.526&81.02 &13-2879-8933-7173-23&28.48&96&1.93\\ \cline{2-7} 
&0.579&79.90 &17-5395-17200-10748-23&49.29&117&2.15\\ \cline{2-7} 
\hline
\end{tabular}
  \begin{tablenotes}
  \item[*] In the order of first hidden layer to last hidden layer out of a total of 288-12288-32768-16384-256  filters, respectively.
  \end{tablenotes}
  \end{threeparttable}
}
\end{subtable}

\vspace{1em}
\begin{subtable}{0.75\textwidth}
\caption{Low-rank CNN model}
\resizebox{\columnwidth}{!}{%
 \begin{threeparttable}
\begin{tabular}{|c|c|c|c|c|c|c|}
\hline
~&$\mu$ &Accuracy ($\%$)&Filter ($\#$)\tnote{*}&Sparsity ($\%$)&Training epochs ($\#$)&Speedup\\ \hline 
 \multirow{8}{*}{\rotatebox{90}{ $l_1$-norm}}
&0&80.20 &0-0-0-0-0-0-0-0-0&0.00&0&1.00\\ \cline{2-7} 
&0.000&81.76 &2-1-2-3-2-2-2-3-2&0.22&4&1.08\\ \cline{2-7} 
&0.105&81.79 &3-2-2-5-5-3-4-3-6&0.64&16&1.21\\ \cline{2-7} 
&0.211&81.75 &6-5-4-5-7-7-6-6-8&0.87&36&1.26\\ \cline{2-7} 
&0.316&81.70 &6-6-8-7-12-15-9-12-24&2.54&56&1.53\\ \cline{2-7} 
&0.421&81.77 &10-11-78-75-221-222-205-208-28&4.09&90&1.69\\ \cline{2-7} 
&0.526&81.36 &14-12-729-728-2241-2246-1804-1806-33&14.83&108&2.18\\ \cline{2-7} 
&0.579&80.14 &18-15-1358-1357-4311-4312-2697-2699-34&23.53&104&2.37\\ \cline{2-7} 
\hline
 \multirow{8}{*}{\rotatebox{90}{ $l_0$-norm}}
&0&80.20 &0-0-0-0-0-0-0-0-0&0.00&0&1.00\\ \cline{2-7} 
&0.000&82.01 &1-3-1-3-1-2-4-1-3&0.33&4&1.11\\ \cline{2-7} 
&0.105&82.01 &5-3-9-9-4-8-9-6-8&0.88&16&1.26\\ \cline{2-7} 
&0.211&81.91 &13-8-9-9-13-12-10-12-13&1.43&30&1.37\\ \cline{2-7} 
&0.316&82.10 &13-9-12-14-20-18-17-19-32&3.41&56&1.62\\ \cline{2-7} 
&0.421&82.00 &14-18-83-87-228-227-212-215-39&5.28&90&1.77\\ \cline{2-7} 
&0.526&81.38 &19-19-737-738-2249-2250-1807-1813-39&15.51&96&2.19\\ \cline{2-7} 
&0.579&80.25 &25-20-1368-1367-4315-4322-2703-2706-40&24.22&104&2.37\\ \cline{2-7} 
\hline
\end{tabular}
  \begin{tablenotes}
  \item[*] In the order of first hidden layer to last hidden layer out of a total of 144-144-6144-6144-16384-16384-8192-8192-256 filters, respectively.
  \end{tablenotes}
  \end{threeparttable}
}
\end{subtable}%
\end{table}
 
\begin{table}[!]
\centering
\caption{Performance of the proposed ADMM-based sparse method on CIFAR-100 dataset.}
\label{tab:CIFAR100}
\begin{subtable}{0.75\textwidth}
\caption{NIN model}
\resizebox{\columnwidth}{!}{%
 \begin{threeparttable}
\begin{tabular}{|c|c|c|c|c|c|c|}
\hline
~&$\mu$ &Accuracy ($\%$)&Filter ($\#$)\tnote{*}&Sparsity ($\%$)&Training epochs ($\#$)&Speedup\\ \hline 
 \multirow{8}{*}{\rotatebox{90}{ $l_1$-norm}}
&0&63.30 &0-0-0-0&0.00&0&1.00\\ \cline{2-7} 
&0.000&63.97 &34-484-970-1&1.24&4&1.07\\ \cline{2-7} 
&0.105&64.33 &36-553-1029-4&1.37&16&1.07\\ \cline{2-7} 
&0.211&64.46 &40-611-1149-35&1.57&30&1.08\\ \cline{2-7} 
&0.316&64.34 &48-751-1433-593&2.74&64&1.14\\ \cline{2-7} 
&0.421&64.18 &65-1824-2307-5639&12.17&90&1.42\\ \cline{2-7} 
&0.526&64.01 &80-6815-4843-12460&30.76&96&1.71\\ \cline{2-7} 
&0.632&64.30 &117-6816-11581-12464&34.12&112&1.75\\ \cline{2-7} 
\hline
 \multirow{8}{*}{\rotatebox{90}{ $l_0$-norm}}
&0&63.30 &0-0-0-0&0.00&0&1.00\\ \cline{2-7} 
&0.000&64.32 &36-485-969-0&1.24&4&1.07\\ \cline{2-7} 
&0.105&64.23 &40-558-1033-6&1.39&16&1.07\\ \cline{2-7} 
&0.211&64.44 &43-618-1152-37&1.59&30&1.08\\ \cline{2-7} 
&0.316&64.76 &50-759-1438-598&2.77&48&1.14\\ \cline{2-7} 
&0.421&64.59 &71-1834-2315-5649&12.21&100&1.42\\ \cline{2-7} 
&0.526&64.10 &85-6824-4850-12472&30.81&120&1.71\\ \cline{2-7} 
&0.632&63.93 &124-6832-11587-12472&34.17&140&1.75\\ \cline{2-7} 
\hline
\end{tabular}
  \begin{tablenotes}
  \item[*] In the order of first hidden layer to last hidden layer out of a total of 576-18432-36864-36864 filters, respectively.
  \end{tablenotes}
  \end{threeparttable}
}
\end{subtable}

\vspace{1em}
\begin{subtable}{0.75\textwidth}
\caption{Low-rank NIN model}
\resizebox{\columnwidth}{!}{%
 \begin{threeparttable}
\begin{tabular}{|c|c|c|c|c|c|c|}
\hline
~&$\mu$ &Accuracy ($\%$)&Filter ($\#$)\tnote{*}&Sparsity ($\%$)&Training epochs ($\#$)&Speedup\\ \hline 
 \multirow{8}{*}{\rotatebox{90}{ $l_1$-norm}}
&0&63.60 &0-0-0-0-0-0-0-0&0.00&0&1.00\\ \cline{2-7} 
&0.000&64.62 &10-9-194-98-102-96-2-3&0.55&4&1.08\\ \cline{2-7} 
&0.105&64.51 &12-11-196-101-183-100-22-9&0.68&16&1.10\\ \cline{2-7} 
&0.211&64.83 &13-14-205-107-291-122-82-16&0.92&36&1.13\\ \cline{2-7} 
&0.316&65.22 &18-18-282-142-489-183-275-60&1.58&48&1.21\\ \cline{2-7} 
&0.421&65.18 &22-31-486-246-1113-417-992-234&3.82&90&1.39\\ \cline{2-7} 
&0.526&65.12 &31-37-1170-653-2843-1349-3092-881&10.84&120&1.72\\ \cline{2-7} 
&0.632&64.85 &42-49-2721-1998-6520-4189-7691-3242&28.52&112&2.08\\ \cline{2-7} 
\hline
 \multirow{8}{*}{\rotatebox{90}{ $l_0$-norm}}
&0&63.60 &0-0-0-0-0-0-0-0&0.00&0&1.00\\ \cline{2-7} 
&0.000&64.58 &8-12-195-96-104-98-3-3&0.56&4&1.08\\ \cline{2-7} 
&0.105&64.90 &15-14-199-103-186-104-24-11&0.71&16&1.10\\ \cline{2-7} 
&0.211&64.93 &21-19-210-110-293-125-88-21&0.96&36&1.14\\ \cline{2-7} 
&0.316&65.07 &25-20-288-148-496-188-281-67&1.63&48&1.21\\ \cline{2-7} 
&0.421&64.93 &30-38-492-253-1118-426-998-240&3.88&80&1.40\\ \cline{2-7} 
&0.526&64.88 &34-44-1181-663-2845-1359-3099-887&10.90&96&1.72\\ \cline{2-7} 
&0.632&65.11 &55-59-2725-2008-6531-4192-7703-3248&28.60&140&2.08\\ \cline{2-7} 
\hline
\end{tabular}
  \begin{tablenotes}
  \item[*] In the order of first hidden layer to last hidden layer out of a total of 288-288-9216-9216-18432-18432-18432-18432 filters, respectively.
  \end{tablenotes}
  \end{threeparttable}
}
\end{subtable}

\vspace{1em}
\begin{subtable}{0.75\textwidth}
\caption{CNN model}
\resizebox{\columnwidth}{!}{%
 \begin{threeparttable}
\begin{tabular}{|c|c|c|c|c|c|c|}
\hline
~&$\mu$ &Accuracy ($\%$)&Filter ($\#$)\tnote{*}&Sparsity ($\%$)&Training epochs ($\#$)&Speedup\\ \hline 
 \multirow{8}{*}{\rotatebox{90}{ $l_1$-norm}}
&0&60.11 &0-0-0-0-0&0.00&0&1.00\\ \cline{2-7} 
&0.000&61.39 &2-1-0-1-2&0.09&4&1.01\\ \cline{2-7} 
&0.105&61.88 &3-3-3-4-2&0.10&16&1.01\\ \cline{2-7} 
&0.211&61.60 &3-3-5-4-4&0.19&30&1.02\\ \cline{2-7} 
&0.316&61.73 &7-11-25-13-23&1.03&64&1.10\\ \cline{2-7} 
&0.421&61.97 &7-274-848-801-23&3.74&80&1.28\\ \cline{2-7} 
&0.526&61.43 &10-2877-8929-7173-23&28.46&96&1.90\\ \cline{2-7} 
&0.579&59.81 &16-5390-17196-10748-23&49.27&117&2.11\\ \cline{2-7} 
\hline
 \multirow{8}{*}{\rotatebox{90}{ $l_0$-norm}}
&0&60.11 &0-0-0-0-0&0.00&0&1.00\\ \cline{2-7} 
&0.000&61.89 &2-3-2-2-3&0.14&4&1.01\\ \cline{2-7} 
&0.105&62.12 &3-3-5-7-6&0.27&16&1.03\\ \cline{2-7} 
&0.211&61.90 &7-5-7-11-6&0.29&36&1.03\\ \cline{2-7} 
&0.316&62.01 &14-15-33-22-27&1.23&56&1.11\\ \cline{2-7} 
&0.421&62.15 &18-285-859-808-29&4.05&90&1.29\\ \cline{2-7} 
&0.526&61.20 &24-2890-8943-7181-32&28.91&120&1.90\\ \cline{2-7} 
&0.579&59.92 &30-5404-17211-10756-35&49.84&104&2.11\\ \cline{2-7} 
\hline
\end{tabular}
  \begin{tablenotes}
  \item[*] In the order of first hidden layer to last hidden layer out of a total of 288-12288-32768-16384-256 filters, respectively.
  \end{tablenotes}
  \end{threeparttable}
}
\end{subtable}

\vspace{1em}
\begin{subtable}{0.75\textwidth}
\caption{Low-rank CNN model}
\resizebox{\columnwidth}{!}{%
 \begin{threeparttable}
\begin{tabular}{|c|c|c|c|c|c|c|}
\hline
~&$\mu$ &Accuracy ($\%$)&Filter ($\#$)\tnote{*}&Sparsity ($\%$)&Training epochs ($\#$)&Speedup\\ \hline 
 \multirow{8}{*}{\rotatebox{90}{ $l_1$-norm}}
&0&60.23 &0-0-0-0-0-0-0-0-0&0.00&0&1.00\\ \cline{2-7} 
&0.000&61.54 &2-2-3-3-3-1-1-2-1&0.12&4&1.04\\ \cline{2-7} 
&0.105&61.98 &3-5-4-4-3-4-4-4-7&0.75&16&1.19\\ \cline{2-7} 
&0.211&61.70 &5-8-4-5-4-5-7-8-9&0.97&30&1.23\\ \cline{2-7} 
&0.316&61.74 &9-8-8-7-11-10-11-12-26&2.75&48&1.47\\ \cline{2-7} 
&0.421&61.96 &10-9-79-74-224-222-206-209-32&4.50&90&1.62\\ \cline{2-7} 
&0.526&61.18 &11-15-730-729-2244-2243-1801-1805-35&15.03&108&2.07\\ \cline{2-7} 
&0.579&60.09 &12-16-1360-1358-4310-4309-2697-2698-35&23.63&104&2.25\\ \cline{2-7} 
\hline
 \multirow{8}{*}{\rotatebox{90}{ $l_0$-norm}}
&0&60.23 &0-0-0-0-0-0-0-0-0&0.00&0&1.00\\ \cline{2-7} 
&0.000&61.62 &1-1-2-1-4-3-2-2-2&0.22&4&1.06\\ \cline{2-7} 
&0.105&62.20 &7-5-8-6-7-6-7-7-8&0.88&16&1.20\\ \cline{2-7} 
&0.211&61.91 &8-8-8-9-10-10-10-8-12&1.31&36&1.28\\ \cline{2-7} 
&0.316&61.94 &15-13-13-11-16-19-19-18-32&3.42&56&1.52\\ \cline{2-7} 
&0.421&62.05 &16-15-82-80-230-233-218-215-36&4.98&90&1.64\\ \cline{2-7} 
&0.526&61.33 &23-16-736-733-2250-2253-1812-1814-42&15.82&120&2.07\\ \cline{2-7} 
&0.579&60.20 &24-19-1365-1364-4316-4319-2706-2707-43&24.52&117&2.25\\ \cline{2-7} 
\hline
\end{tabular}
  \begin{tablenotes}
  \item[*] In the order of first hidden layer to last hidden layer out of a total of 144-144-6144-6144-16384-16384-8192-8192-256 filters, respectively.
  \end{tablenotes}
  \end{threeparttable}
}
\end{subtable}%
\end{table}

\begin{table}[!htbp]
\centering
\caption{Performance of the proposed ADMM-based sparse method on SVHN dataset.}
\label{tab:SVHN}
\begin{subtable}{0.75\textwidth}
\caption{NIN model}
\resizebox{\columnwidth}{!}{%
 \begin{threeparttable}
\begin{tabular}{|c|c|c|c|c|c|c|}
\hline
~&$\mu$ &Accuracy ($\%$)&Filter ($\#$)\tnote{*}&Sparsity ($\%$)&Training epochs ($\#$)&Speedup\\ \hline 
 \multirow{8}{*}{\rotatebox{90}{ $l_1$-norm}}
&0&96.20 &0-0-0-0&0.00&0&1.00\\ \cline{2-7} 
&0.000&96.90 &34-483-970-0&1.23&4&1.09\\ \cline{2-7} 
&0.105&97.02 &36-554-1031-3&1.37&16&1.10\\ \cline{2-7} 
&0.211&97.32 &38-612-1148-34&1.56&30&1.11\\ \cline{2-7} 
&0.316&97.36 &45-752-1432-592&2.74&48&1.18\\ \cline{2-7} 
&0.421&97.06 &61-1827-2304-5640&12.17&90&1.52\\ \cline{2-7} 
&0.526&96.66 &81-6815-4838-12461&30.77&96&1.84\\ \cline{2-7} 
&0.632&96.96 &118-6816-11581-12461&34.12&126&1.88\\ \cline{2-7} 
\hline
 \multirow{8}{*}{\rotatebox{90}{ $l_0$-norm}}
&0&96.20 &0-0-0-0&0.00&0&1.00\\ \cline{2-7} 
&0.000&96.89 &35-485-969-1&1.24&4&1.09\\ \cline{2-7} 
&0.105&97.23 &38-559-1031-5&1.39&16&1.10\\ \cline{2-7} 
&0.211&97.47 &41-618-1149-40&1.59&36&1.12\\ \cline{2-7} 
&0.316&97.49 &52-764-1436-598&2.78&56&1.19\\ \cline{2-7} 
&0.421&97.56 &67-1837-2314-5643&12.20&90&1.53\\ \cline{2-7} 
&0.526&97.16 &93-6827-4851-12464&30.81&120&1.84\\ \cline{2-7} 
&0.632&97.07 &130-6829-11586-12470&34.17&126&1.88\\ \cline{2-7} 
\hline
\end{tabular}
  \begin{tablenotes}
  \item[*] In the order of first hidden layer to last hidden layer out of a total of 576-18432-36864-36864 filters, respectively.
  \end{tablenotes}
  \end{threeparttable}
}
\end{subtable}

\vspace{1em}
\begin{subtable}{0.75\textwidth}
\caption{Low-rank NIN model}
\resizebox{\columnwidth}{!}{%
 \begin{threeparttable}
\begin{tabular}{|c|c|c|c|c|c|c|}
\hline
~&$\mu$ &Accuracy ($\%$)&Filter ($\#$)\tnote{*}&Sparsity ($\%$)&Training epochs ($\#$)&Speedup\\ \hline 
 \multirow{8}{*}{\rotatebox{90}{ $l_1$-norm}}
&0&96.70 &0-0-0-0-0-0-0-0&0.00&0&1.00\\ \cline{2-7} 
&0.000&97.72 &10-11-193-98-102-98-0-3&0.56&4&1.11\\ \cline{2-7} 
&0.105&97.58 &11-12-194-102-182-99-21-10&0.68&16&1.13\\ \cline{2-7} 
&0.211&98.16 &15-15-206-108-292-121-80-18&0.92&36&1.16\\ \cline{2-7} 
&0.316&97.90 &17-16-280-139-490-182-276-65&1.58&64&1.25\\ \cline{2-7} 
&0.421&98.18 &21-31-485-247-1112-416-993-232&3.81&100&1.46\\ \cline{2-7} 
&0.526&97.98 &27-37-1174-652-2840-1348-3093-882&10.84&120&1.80\\ \cline{2-7} 
&0.632&98.11 &41-46-2718-2002-6517-4189-7694-3243&28.52&126&2.18\\ \cline{2-7} 
\hline
 \multirow{8}{*}{\rotatebox{90}{ $l_0$-norm}}
&0&96.70 &0-0-0-0-0-0-0-0&0.00&0&1.00\\ \cline{2-7} 
&0.000&97.51 &8-9-195-96-103-99-3-5&0.56&4&1.11\\ \cline{2-7} 
&0.105&97.88 &16-13-196-104-185-103-26-11&0.71&16&1.13\\ \cline{2-7} 
&0.211&97.94 &17-18-209-113-297-126-86-21&0.96&36&1.17\\ \cline{2-7} 
&0.316&98.27 &25-24-287-144-494-187-281-71&1.63&64&1.26\\ \cline{2-7} 
&0.421&97.97 &30-36-492-251-1119-423-997-239&3.87&100&1.46\\ \cline{2-7} 
&0.526&98.45 &39-42-1176-659-2849-1363-3102-891&10.91&96&1.81\\ \cline{2-7} 
&0.632&97.99 &53-57-2725-2004-6535-4201-7705-3257&28.62&112&2.18\\ \cline{2-7} 
\hline
\end{tabular}
  \begin{tablenotes}
  \item[*] In the order of first hidden layer to last hidden layer out of a total of 288-288-9216-9216-18432-18432-18432-18432 filters, respectively.
  \end{tablenotes}
  \end{threeparttable}
}
\end{subtable}

\vspace{1em}
\begin{subtable}{0.75\textwidth}
\caption{CNN model}
\resizebox{\columnwidth}{!}{%
 \begin{threeparttable}
\begin{tabular}{|c|c|c|c|c|c|c|}
\hline
~&$\mu$ &Accuracy ($\%$)&Filter ($\#$)\tnote{*}&Sparsity ($\%$)&Training epochs ($\#$)&Speedup\\ \hline 
 \multirow{8}{*}{\rotatebox{90}{ $l_1$-norm}}
&0&85.10 &0-0-0-0-0&0.00&0&1.00\\ \cline{2-7} 
&0.000&86.81 &2-1-0-2-0&0.01&4&1.00\\ \cline{2-7} 
&0.105&86.68 &2-3-2-2-4&0.18&16&1.02\\ \cline{2-7} 
&0.211&86.69 &5-5-6-3-4&0.19&36&1.02\\ \cline{2-7} 
&0.316&86.35 &6-10-28-14-19&0.87&64&1.10\\ \cline{2-7} 
&0.421&86.85 &6-277-853-798-24&3.79&80&1.32\\ \cline{2-7} 
&0.526&86.34 &9-2880-8932-7167-24&28.50&96&1.97\\ \cline{2-7} 
&0.579&83.30 &13-5393-17199-10748-25&49.36&104&2.19\\ \cline{2-7} 
\hline
 \multirow{8}{*}{\rotatebox{90}{ $l_0$-norm}}
&0&85.10 &0-0-0-0-0&0.00&0&1.00\\ \cline{2-7} 
&0.000&86.63 &1-0-1-0-2&0.09&4&1.01\\ \cline{2-7} 
&0.105&86.70 &4-4-5-7-6&0.28&16&1.03\\ \cline{2-7} 
&0.211&86.80 &11-8-9-8-7&0.34&30&1.04\\ \cline{2-7} 
&0.316&86.74 &13-17-32-17-22&1.02&64&1.11\\ \cline{2-7} 
&0.421&86.97 &13-285-855-809-29&4.04&90&1.34\\ \cline{2-7} 
&0.526&86.49 &19-2888-8934-7178-29&28.76&120&1.97\\ \cline{2-7} 
&0.579&83.40 &26-5401-17206-10758-29&49.58&117&2.19\\ \cline{2-7} 
\hline
\end{tabular}
  \begin{tablenotes}
  \item[*] In the order of first hidden layer to last hidden layer out of a total of 288-12288-32768-16384-256 filters, respectively.
  \end{tablenotes}
  \end{threeparttable}
}
\end{subtable}

\vspace{1em}
\begin{subtable}{0.75\textwidth}
\caption{Low-rank CNN model}
\resizebox{\columnwidth}{!}{%
 \begin{threeparttable}
\begin{tabular}{|c|c|c|c|c|c|c|}
\hline
~&$\mu$ &Accuracy ($\%$)&Filter ($\#$)\tnote{*}&Sparsity ($\%$)&Training epochs ($\#$)&Speedup\\ \hline 
 \multirow{8}{*}{\rotatebox{90}{ $l_1$-norm}}
&0&87.60 &0-0-0-0-0-0-0-0-0&0.00&0&1.00\\ \cline{2-7} 
&0.000&88.93 &3-3-3-1-3-2-3-3-1&0.13&4&1.04\\ \cline{2-7} 
&0.105&89.28 &3-4-5-2-5-5-3-4-3&0.34&16&1.10\\ \cline{2-7} 
&0.211&89.39 &7-5-7-7-7-6-5-5-6&0.67&36&1.18\\ \cline{2-7} 
&0.316&89.18 &10-7-9-10-12-11-10-9-25&2.65&48&1.49\\ \cline{2-7} 
&0.421&89.55 &11-11-76-77-222-219-209-208-33&4.61&100&1.66\\ \cline{2-7} 
&0.526&88.83 &14-14-727-732-2242-2242-1803-1802-33&14.83&96&2.10\\ \cline{2-7} 
&0.579&86.30 &15-15-1357-1361-4308-4312-2696-2695-36&23.73&130&2.29\\ \cline{2-7} 
\hline
 \multirow{8}{*}{\rotatebox{90}{ $l_0$-norm}}
&0&87.60 &0-0-0-0-0-0-0-0-0&0.00&0&1.00\\ \cline{2-7} 
&0.000&89.30 &1-5-5-4-4-1-3-3-1&0.13&4&1.04\\ \cline{2-7} 
&0.105&89.55 &4-9-5-7-6-7-5-5-7&0.77&16&1.20\\ \cline{2-7} 
&0.211&89.65 &6-11-9-9-12-12-6-11-11&1.21&30&1.28\\ \cline{2-7} 
&0.316&89.39 &10-15-11-14-17-18-17-18-29&3.10&56&1.52\\ \cline{2-7} 
&0.421&89.56 &17-16-83-85-224-232-213-216-42&5.59&90&1.72\\ \cline{2-7} 
&0.526&89.00 &21-23-740-738-2247-2255-1813-1813-44&16.04&120&2.12\\ \cline{2-7} 
&0.579&86.41 &25-24-1369-1367-4316-4321-2706-2706-44&24.64&117&2.29\\ \cline{2-7} 
\hline
\end{tabular}
  \begin{tablenotes}
  \item[*] In the order of first hidden layer to last hidden layer out of a total of 144-144-6144-6144-16384-16384-8192-8192-256 filters, respectively.
  \end{tablenotes}
  \end{threeparttable}
}
\end{subtable}%
\end{table}

\begin{table}[!htbp]
\centering
\caption{Joint variations of accuracy and sparsity on the evaluated datasets for increasing $\mu$ values. LR-NIN stands for low-rank NIN while LR-CNN is for low-rank CNN.}
\label{tab:ResultsCombined}
\begin{subtable}{0.525\textwidth}
\caption{CIFAR-10}
\label{tab:ResultsCombinedCIFAR10}
\resizebox{\columnwidth}{!}{%
\begin{tabular}{|c|c|c|c|c|c|c|c|c|c|c|}
\hline
 \multirow{8}{*}{\rotatebox{90}{ NIN}}
 &\multirow{4}{*}{\rotatebox{90}{ $l_1$}}
&Sparsity ($\%$)&0.00&1.23&1.36&1.55&2.72&12.14&30.73&34.07\\ \cline{3-11} 
&&Accuracy ($\%$)&90.71&91.14&91.42&91.47&91.56&91.47&91.13&91.21\\ \cline{3-11} 
&&Training epochs ($\#$)&0&4&16&30&48&90&120&140\\ \cline{3-11} 
&&Speedup&1.00&1.08&1.09&1.10&1.16&1.47&1.77&1.81\\ \cline{2-11} 
& \multirow{4}{*}{\rotatebox{90}{ $l_0$}}
&Sparsity ($\%$)&0.00&1.23&1.37&1.57&2.75&12.18&30.78&34.12\\ \cline{3-11} 
&&Accuracy ($\%$)&90.71&91.24&91.52&91.57&91.66&91.57&91.23&91.31\\ \cline{3-11} 
&&Training epochs ($\#$)&0&4&16&36&64&80&96&112\\ \cline{3-11} 
&&Speedup&1.00&1.08&1.09&1.10&1.16&1.47&1.77&1.81\\ \cline{2-11} 
\hline
 \multirow{8}{*}{\rotatebox{90}{ LR-NIN }}
&\multirow{4}{*}{\rotatebox{90}{ $l_1$}}
&Sparsity ($\%$)&0.00&0.54&0.66&0.88&1.53&3.75&10.76&28.43\\ \cline{3-11} 
&&Accuracy ($\%$)&90.07&90.65&90.92&91.12&91.25&91.22&91.21&91.20\\ \cline{3-11} 
&&Training epochs ($\#$)&0&4&16&30&56&100&120&140\\ \cline{3-11} 
&&Speedup&1.00&1.09&1.11&1.14&1.22&1.42&1.76&2.13\\ \cline{2-11} 
& \multirow{4}{*}{\rotatebox{90}{ $l_0$}}
&Sparsity ($\%$)&0.00&0.55&0.68&0.91&1.58&3.82&10.84&28.51\\ \cline{3-11} 
&&Accuracy ($\%$)&90.07&90.75&91.02&91.22&91.35&91.32&91.31&91.30\\ \cline{3-11} 
&&Training epochs ($\#$)&0&4&16&36&48&80&108&126\\ \cline{3-11} 
&&Speedup&1.00&1.09&1.11&1.15&1.23&1.43&1.76&2.13\\ \cline{2-11} 
\hline
 \multirow{8}{*}{\rotatebox{90}{ CNN}}
&\multirow{4}{*}{\rotatebox{90}{ $l_1$}}
&Sparsity ($\%$)&0.00&0.00&0.00&0.00&2.40&5.11&29.82&50.63\\ \cline{3-11} 
&&Accuracy ($\%$)&80.00&81.24&81.44&81.46&81.24&81.48&80.92&79.80\\ \cline{3-11} 
&&Training epochs ($\#$)&0&4&16&30&64&80&120&130\\ \cline{3-11} 
&&Speedup&1.00&1.00&1.00&1.00&1.21&1.37&1.94&2.15\\ \cline{2-11} 
&\multirow{4}{*}{\rotatebox{90}{ $l_0$}}
&Sparsity ($\%$)&0.00&0.05&0.14&0.23&0.95&3.75&28.48&49.29\\ \cline{3-11} 
&&Accuracy ($\%$)&80.00&81.34&81.54&81.56&81.34&81.58&81.02&79.90\\ \cline{3-11} 
&&Training epochs ($\#$)&0&4&16&36&56&80&96&117\\ \cline{3-11} 
&&Speedup&1.00&1.01&1.02&1.03&1.10&1.30&1.93&2.15\\ \cline{2-11} 
\hline
 \multirow{8}{*}{\rotatebox{90}{ LR-CNN}}
&\multirow{4}{*}{\rotatebox{90}{ $l_1$}}
&Sparsity ($\%$)&0.00&0.22&0.64&0.87&2.54&4.09&14.83&23.53\\ \cline{3-11} 
&&Accuracy ($\%$)&80.20&81.76&81.79&81.75&81.70&81.77&81.36&80.14\\ \cline{3-11} 
&&Training epochs ($\#$)&0&4&16&36&56&90&108&104\\ \cline{3-11} 
&&Speedup&1.00&1.08&1.21&1.26&1.53&1.69&2.18&2.37\\ \cline{2-11} 
&\multirow{4}{*}{\rotatebox{90}{ $l_0$}}
&Sparsity ($\%$)&0.00&0.33&0.88&1.43&3.41&5.28&15.51&24.22\\ \cline{3-11} 
&&Accuracy ($\%$)&80.20&82.01&82.01&81.91&82.10&82.00&81.38&80.25\\ \cline{3-11} 
&&Training epochs ($\#$)&0&4&16&30&56&90&96&104\\ \cline{3-11} 
&&Speedup&1.00&1.11&1.26&1.37&1.62&1.77&2.19&2.37\\ \cline{2-11} 
\hline
\end{tabular}
}
\end{subtable}

\vspace{1em}
\begin{subtable}{0.525\textwidth}
\caption{CIFAR-100}
\label{tab:ResultsCombinedCIFAR100}
\resizebox{\columnwidth}{!}{%
\begin{tabular}{|c|c|c|c|c|c|c|c|c|c|c|}
\hline
 \multirow{8}{*}{\rotatebox{90}{ NIN}}
 &\multirow{4}{*}{\rotatebox{90}{ $l_1$}}
&Sparsity ($\%$)&0.00&1.24&1.37&1.57&2.74&12.17&30.76&34.12\\ \cline{3-11} 
&&Accuracy ($\%$)&63.30&63.97&64.33&64.46&64.34&64.18&64.01&64.30\\ \cline{3-11} 
&&Training epochs ($\#$)&0&4&16&30&64&90&96&112\\ \cline{3-11} 
&&Speedup&1.00&1.07&1.07&1.08&1.14&1.42&1.71&1.75\\ \cline{2-11} 
& \multirow{4}{*}{\rotatebox{90}{ $l_0$}}
&Sparsity ($\%$)&0.00&1.24&1.39&1.59&2.77&12.21&30.81&34.17\\ \cline{3-11} 
&&Accuracy ($\%$)&63.30&64.32&64.23&64.44&64.76&64.59&64.10&63.93\\ \cline{3-11} 
&&Training epochs ($\#$)&0&4&16&30&48&100&120&140\\ \cline{3-11} 
&&Speedup&1.00&1.07&1.07&1.08&1.14&1.42&1.71&1.75\\ \cline{2-11} 
\hline
 \multirow{8}{*}{\rotatebox{90}{ LR-NIN }}
&\multirow{4}{*}{\rotatebox{90}{ $l_1$}}
&Sparsity ($\%$)&0.00&0.55&0.68&0.92&1.58&3.82&10.84&28.52\\ \cline{3-11} 
&&Accuracy ($\%$)&63.60&64.62&64.51&64.83&65.22&65.18&65.12&64.85\\ \cline{3-11} 
&&Training epochs ($\#$)&0&4&16&36&48&90&120&112\\ \cline{3-11} 
&&Speedup&1.00&1.08&1.10&1.13&1.21&1.39&1.72&2.08\\ \cline{2-11} 
& \multirow{4}{*}{\rotatebox{90}{ $l_0$}}
&Sparsity ($\%$)&0.00&0.56&0.71&0.96&1.63&3.88&10.90&28.60\\ \cline{3-11} 
&&Accuracy ($\%$)&63.60&64.58&64.90&64.93&65.07&64.93&64.88&65.11\\ \cline{3-11} 
&&Training epochs ($\#$)&0&4&16&36&48&80&96&140\\ \cline{3-11} 
&&Speedup&1.00&1.08&1.10&1.14&1.21&1.40&1.72&2.08\\ \cline{2-11} 
\hline
 \multirow{8}{*}{\rotatebox{90}{ CNN}}
&\multirow{4}{*}{\rotatebox{90}{ $l_1$}}
&Sparsity ($\%$)&0.00&0.09&0.10&0.19&1.03&3.74&28.46&49.27\\ \cline{3-11} 
&&Accuracy ($\%$)&60.11&61.39&61.88&61.60&61.73&61.97&61.43&59.81\\ \cline{3-11} 
&&Training epochs ($\#$)&0&4&16&30&64&80&96&117\\ \cline{3-11} 
&&Speedup&1.00&1.01&1.01&1.02&1.10&1.28&1.90&2.11\\ \cline{2-11} 
&\multirow{4}{*}{\rotatebox{90}{ $l_0$}}
&Sparsity ($\%$)&0.00&0.14&0.27&0.29&1.23&4.05&28.91&49.84\\ \cline{3-11} 
&&Accuracy ($\%$)&60.11&61.89&62.12&61.90&62.01&62.15&61.20&59.92\\ \cline{3-11} 
&&Training epochs ($\#$)&0&4&16&36&56&90&120&104\\ \cline{3-11} 
&&Speedup&1.00&1.01&1.03&1.03&1.11&1.29&1.90&2.11\\ \cline{2-11} 
\hline
 \multirow{8}{*}{\rotatebox{90}{ LR-CNN}}
&\multirow{4}{*}{\rotatebox{90}{ $l_1$}} 
&Sparsity ($\%$)&0.00&0.12&0.75&0.97&2.75&4.50&15.03&23.63\\ \cline{3-11} 
&&Accuracy ($\%$)&60.23&61.54&61.98&61.70&61.74&61.96&61.18&60.09\\ \cline{3-11} 
&&Training epochs ($\#$)&0&4&16&30&48&90&108&104\\ \cline{3-11} 
&&Speedup&1.00&1.04&1.19&1.23&1.47&1.62&2.07&2.25\\ \cline{2-11} 
&\multirow{4}{*}{\rotatebox{90}{ $l_0$}} 
&Sparsity ($\%$)&0.00&0.22&0.88&1.31&3.42&4.98&15.82&24.52\\ \cline{3-11} 
&&Accuracy ($\%$)&60.23&61.62&62.20&61.91&61.94&62.05&61.33&60.20\\ \cline{3-11} 
&&Training epochs ($\#$)&0&4&16&36&56&90&120&117\\ \cline{3-11} 
&&Speedup&1.00&1.06&1.20&1.28&1.52&1.64&2.07&2.25\\ \cline{2-11} 
\hline
\end{tabular}
}
\end{subtable}

\vspace{1em}
\begin{subtable}{0.525\textwidth}
\caption{SVHN}
\label{tab:ResultsCombinedSVHN}
\resizebox{\columnwidth}{!}{%
\begin{tabular}{|c|c|c|c|c|c|c|c|c|c|c|}
\hline
 \multirow{8}{*}{\rotatebox{90}{ NIN}}
 &\multirow{4}{*}{\rotatebox{90}{ $l_1$}}
&Sparsity ($\%$)&0.00&1.23&1.37&1.56&2.74&12.17&30.77&34.12\\ \cline{3-11} 
&&Accuracy ($\%$)&96.20&96.90&97.02&97.32&97.36&97.06&96.66&96.96\\ \cline{3-11} 
&&Training epochs ($\#$)&0&4&16&30&48&90&96&126\\ \cline{3-11} 
&&Speedup&1.00&1.09&1.10&1.11&1.18&1.52&1.84&1.88\\ \cline{2-11} 
& \multirow{4}{*}{\rotatebox{90}{ $l_0$}}
&Sparsity ($\%$)&0.00&1.24&1.39&1.59&2.78&12.20&30.81&34.17\\ \cline{3-11} 
&&Accuracy ($\%$)&96.20&96.89&97.23&97.47&97.49&97.56&97.16&97.07\\ \cline{3-11} 
&&Training epochs ($\#$)&0&4&16&36&56&90&120&126\\ \cline{3-11} 
&&Speedup&1.00&1.09&1.10&1.12&1.19&1.53&1.84&1.88\\ \cline{2-11} 
\hline
 \multirow{8}{*}{\rotatebox{90}{ LR-NIN }}
&\multirow{4}{*}{\rotatebox{90}{ $l_1$}}
&Sparsity ($\%$)&0.00&0.56&0.68&0.92&1.58&3.81&10.84&28.52\\ \cline{3-11} 
&&Accuracy ($\%$)&96.70&97.72&97.58&98.16&97.90&98.18&97.98&98.11\\ \cline{3-11} 
&&Training epochs ($\#$)&0&4&16&36&64&100&120&126\\ \cline{3-11} 
&&Speedup&1.00&1.11&1.13&1.16&1.25&1.46&1.80&2.18\\ \cline{2-11} 
& \multirow{4}{*}{\rotatebox{90}{ $l_0$}}
&Sparsity ($\%$)&0.00&0.56&0.71&0.96&1.63&3.87&10.91&28.62\\ \cline{3-11} 
&&Accuracy ($\%$)&96.70&97.51&97.88&97.94&98.27&97.97&98.45&97.99\\ \cline{3-11} 
&&Training epochs ($\#$)&0&4&16&36&64&100&96&112\\ \cline{3-11} 
&&Speedup&1.00&1.11&1.13&1.17&1.26&1.46&1.81&2.18\\ \cline{2-11} 
\hline
 \multirow{8}{*}{\rotatebox{90}{ CNN}}
&\multirow{4}{*}{\rotatebox{90}{ $l_1$}}
&Sparsity ($\%$)&0.00&0.01&0.18&0.19&0.87&3.79&28.50&49.36\\ \cline{3-11} 
&&Accuracy ($\%$)&85.10&86.81&86.68&86.69&86.35&86.85&86.34&83.30\\ \cline{3-11} 
&&Training epochs ($\#$)&0&4&16&36&64&80&96&104\\ \cline{3-11} 
&&Speedup&1.00&1.00&1.02&1.02&1.10&1.32&1.97&2.19\\ \cline{2-11} 
&\multirow{4}{*}{\rotatebox{90}{ $l_0$}}
&Sparsity ($\%$)&0.00&0.09&0.28&0.34&1.02&4.04&28.76&49.58\\ \cline{3-11} 
&&Accuracy ($\%$)&85.10&86.63&86.70&86.80&86.74&86.97&86.49&83.40\\ \cline{3-11} 
&&Training epochs ($\#$)&0&4&16&30&64&90&120&117\\ \cline{3-11} 
&&Speedup&1.00&1.01&1.03&1.04&1.11&1.34&1.97&2.19\\ \cline{2-11} 
\hline
 \multirow{8}{*}{\rotatebox{90}{ LR-CNN}}
&\multirow{4}{*}{\rotatebox{90}{ $l_1$}} 
&Sparsity ($\%$)&0.00&0.13&0.34&0.67&2.65&4.61&14.83&23.73\\ \cline{3-11} 
&&Accuracy ($\%$)&87.60&88.93&89.28&89.39&89.18&89.55&88.83&86.30\\ \cline{3-11} 
&&Training epochs ($\#$)&0&4&16&36&48&100&96&130\\ \cline{3-11} 
&&Speedup&1.00&1.04&1.10&1.18&1.49&1.66&2.10&2.29\\ \cline{2-11} 
&\multirow{4}{*}{\rotatebox{90}{ $l_0$}} 
&Sparsity ($\%$)&0.00&0.13&0.77&1.21&3.10&5.59&16.04&24.64\\ \cline{3-11} 
&&Accuracy ($\%$)&87.60&89.30&89.55&89.65&89.39&89.56&89.00&86.41\\ \cline{3-11} 
&&Training epochs ($\#$)&0&4&16&30&56&90&120&117\\ \cline{3-11} 
&&Speedup&1.00&1.04&1.20&1.28&1.52&1.72&2.12&2.29\\ \cline{2-11} 
\end{tabular}
}
\end{subtable}
\end{table}

\end{appendices}
\end{document}